\documentclass[letterpaper]{article} 
\usepackage{aaai25}  
\usepackage{times}  
\usepackage{helvet}  
\usepackage{courier}  
\usepackage[hyphens]{url}  
\usepackage{graphicx} 
\usepackage{natbib}  
\usepackage{caption} 
\usepackage{algorithm}
\usepackage{algorithmic}
\usepackage{subfigure}
\usepackage{multirow}
\usepackage{amsmath}
\usepackage{newfloat}
\usepackage{listings}
\DeclareCaptionStyle{ruled}{labelfont=normalfont,labelsep=colon,strut=off} 
\floatstyle{ruled}
\newfloat{listing}{tb}{lst}{}
\floatname{listing}{Listing}
\title{One Small and One Large for Document-level Event Argument Extraction}
\author {
    Jiaren Peng\textsuperscript{\rm 1}\equalcontrib,
    Hongda Sun\textsuperscript{\rm 2}\equalcontrib,
    Wenzhong Yang\textsuperscript{\rm 1} \footnote{corresponding author},
    Fuyuan Wei\textsuperscript{\rm 1},
    Liang He\textsuperscript{\rm 3},
    Liejun Wang\textsuperscript{\rm 1},
}
\affiliations {
    \textsuperscript{\rm 1}School of Computer Science and Technology (School of Cyberspace Security), Xinjiang University, China\\
    \textsuperscript{\rm 2}Gaoling School of Artificial Intelligence, Renmin University, China\\
    \textsuperscript{\rm 3}Department of Electronic Engineering, Tsinghua University, China\\
    1354527247@qq.com, sunhongda98@ruc.edu.cn, yangwenzhong@xju.edu.cn, wfy@stu.xju.edu.cn, heliang@tsinghua.edu.cn, wljxju@xju.edu.cn
}

\usepackage{bibentry}

\begin{document}

\maketitle

\begin{abstract}

Document-level Event Argument Extraction (EAE) faces two challenges due to increased input length: 1) difficulty in distinguishing semantic boundaries between events, and 2) interference from redundant information.
To address these issues, we propose two methods. The first method introduces the Co and Structure Event Argument Extraction model (CsEAE) based on Small Language Models (SLMs). 
CsEAE includes a co-occurrences-aware module, which integrates information about all events present in the current input through context labeling and co-occurrences event prompts extraction. Additionally, CsEAE includes a structure-aware module that reduces interference from redundant information by establishing structural relationships between the sentence containing the trigger and other sentences in the document.
The second method introduces new prompts to transform the extraction task into a generative task suitable for Large Language Models (LLMs), addressing gaps in EAE performance using LLMs under Supervised Fine-Tuning (SFT) conditions. We also fine-tuned multiple datasets to develop an LLM that performs better across most datasets. Finally, we applied insights from CsEAE to LLMs, achieving further performance improvements. This suggests that reliable insights validated on SLMs are also applicable to LLMs.
We tested our models on the Rams, WikiEvents, and MLEE datasets. The CsEAE model achieved improvements of 2.1\%, 2.3\%, and 3.2\% in the Arg-C F1 metric compared to the baseline, PAIE~\cite{PAIE}. For LLMs, we demonstrated that their performance on document-level datasets is comparable to that of SLMs~\footnote{All code is available at https://github.com/simon-p-j-r/CsEAE}.
\end{abstract}

\section{Introduction}

\begin{figure}
    \centering
    \includegraphics[width=0.9\columnwidth]{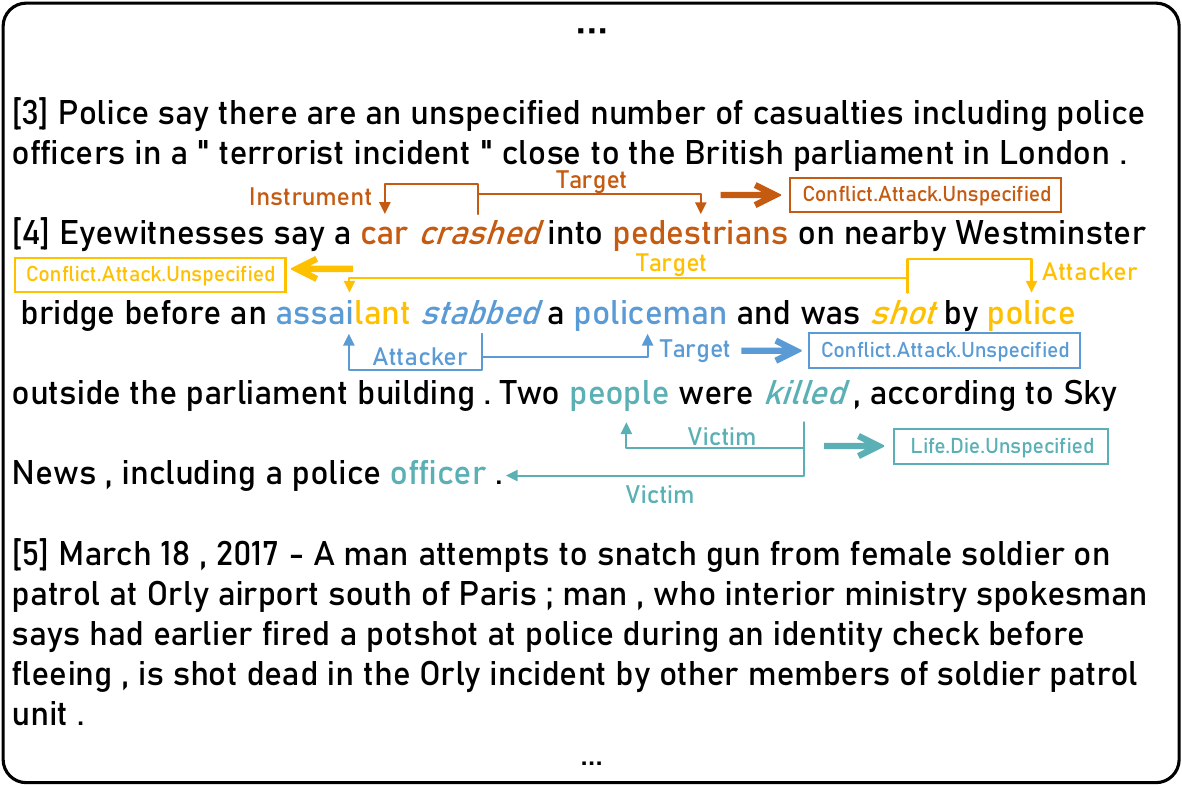}
    \caption{An EAE instance from the WikiEvents dataset.}
    \label{fig:intro}
\end{figure}

Event Argument Extraction (EAE) aims to extract structured event information composed of arguments corresponding to event roles from text \cite{jiaren}. As shown in Figure \ref{fig:intro}, given a trigger and event type, along with a predefined list of roles for the event type, the model needs to extract the corresponding token spans as arguments for each role. This structured information can enhance the performance of downstream tasks such as question answering \cite{question-anwser}, dialogue systems \cite{dialog-system}, and recommendation systems \cite{recommendation}. 

As the length of document-level input texts increases, document-level EAE faces two critical challenges: (1) difficulty in distinguishing semantic boundaries between events \cite{TabEAE}. As shown in Figure \ref{fig:intro}, the four trigger words \textit{crashed}, \textit{stabbed}, \textit{shot}, and \textit{killed}, each trigger four events. The argument distribution of these events is extremely dense, and different events can share the same token span as arguments corresponding to different roles. These dense and overlapping events make the semantic boundaries between them blurry.
(2) The volume of information received by the model increases significantly; however, this information includes not only useful data for the extraction task but also a large amount of redundant information that interferes with the task \cite{TSAR}. For example, in the sentence [5], the presence of person nouns such as \textit{man}, \textit{female} and \textit{soldier} can mislead the extraction of the \textit{Victim} role for the \textit{Life.Die.Unspecified} event triggered by \textit{killed}. However, previous work has not simultaneously addressed both of these issues \cite{PAIE,TSAR,TabEAE}.

\begin{figure*}[!ht]
    \centering
    \includegraphics[width=0.85\textwidth]{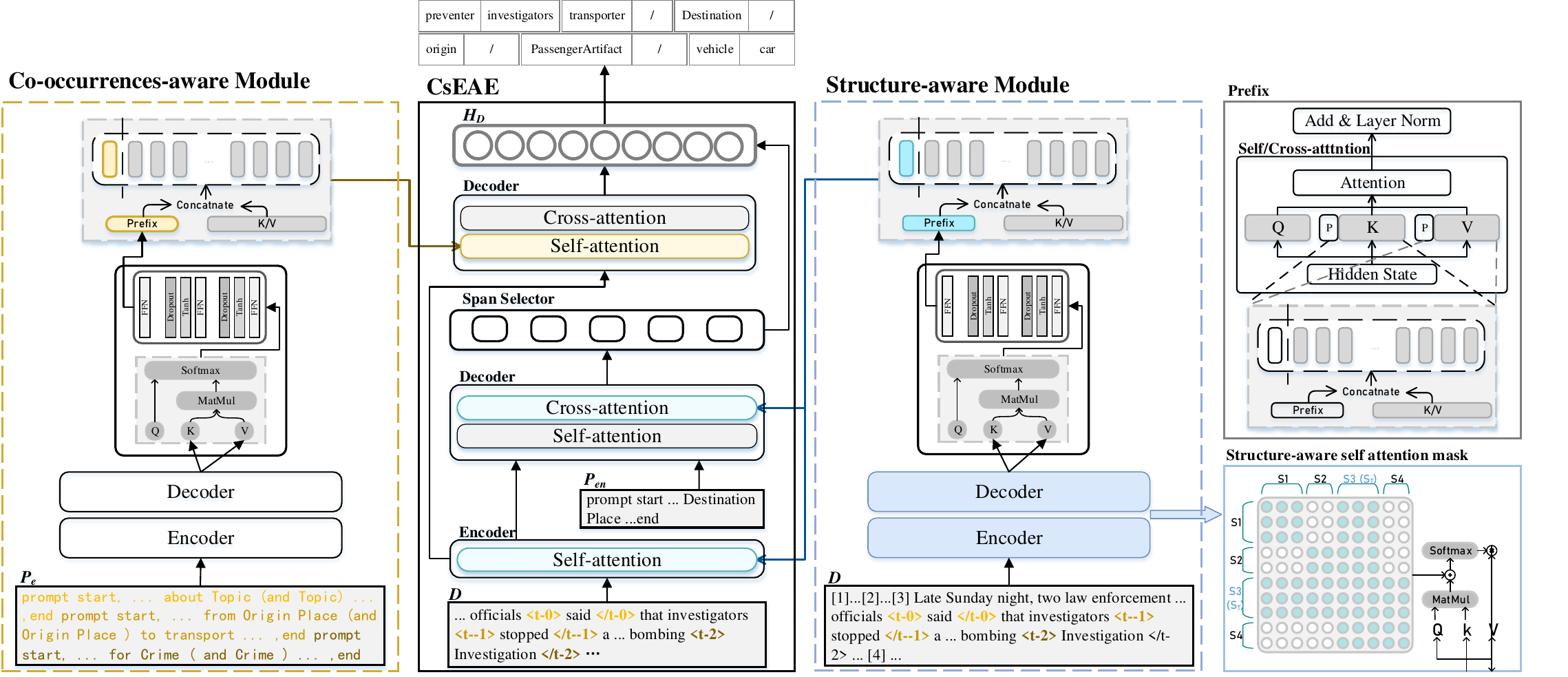}
    \caption{Overview of CsEAE. The yellow attention represents the concatenation of co-occurrences-aware module, while the blue attention represents the concatenation of structure-aware module.}
    \label{fig:prefix}
\end{figure*}

To address these issues, we proposed two methods, with the first being the co and structure EAE model (CsEAE) based on Small Language Models (SLMs). CsEAE enhances the boundaries of the model's focus from both event and sentence perspectives.
From the event perspective, to help the model capture semantic boundaries between events, we introduced a co-occurrence-aware module. This module identifies all co-occurring events in the input by marking triggers and encoding related prompts.
From the sentence perspective, while event mentions are document-level, event information is often within a single sentence. For instance, in the WikiEvents dataset, over 94\% of arguments are in the same sentence as the trigger; in the Rams dataset, over 82\%; and in the MLEE dataset, over 99\%. This highlights the importance of the information in the trigger sentence for the extraction task.
To emphasize this, we structured the knowledge around the trigger sentence and its relationship with other sentences in the document. This approach helps the model selectively gather relevant information from other sentences, reducing distractions from redundant information.

Additionally, we proposed a second method based on Large Language Models (LLMs). We designed prompts tailored to LLMs for each dataset and performed Supervised Fine-Tuning (SFT) on the LLMs. This approach addresses a gap in the EAE field, which previously lacked fine-tuned LLMs \cite{LLM-IE,LLM-EE-Annotator,LLM-EAE}.
Inspired by the use of large-scale high-quality data for continuous pretraining \cite{zhongjing}, we attempted multi-dataset fine-tuning to make the LLMs more familiar with event extraction tasks. On this basis, we also conducted enhanced training on the LLMs using additional datasets.




Finally, inspired by CsEAE, where co-occurrence- and structure-aware interactions enhance the model's ability to capture event boundaries and reduce interference from redundant information, we applied these insights to LLMs. This led to further performance improvements and introduced a novel perspective: the reliable insights validated on SLMs are also applicable to LLMs.Our contributions are summarized below:

$\bullet$ We propose the CsEAE model, which incorporates a co-occurrences-aware module to capture semantic boundaries between events. Additionally, it uses a structure-aware module to build structured perception information, allowing the model to minimize interference from redundant information.


$\bullet$ We designed different prompts for various datasets and further used SFT to enhance the performance of LLMs. Additionally, we proposed multiple datasets SFT and supplementary dataset enhancement training, which led to even better performance.

$\bullet$ We applied insights from SLMs to LLMs, resulting in further performance improvements. This shows that reliable insights validated on SLMs are also effective for LLMs.



\section{Related Works}

\subsection{Document-level Event Argument Extraction}

With the capability to extract events across multiple sentences, document-level EAE has garnered significant research interest. Some studies incorporate abstract meaning representation into the extraction task \cite{TSAR, TARA, AMPERE}. BART-Gen \cite{DOC_EE} utilizes a prompt-based generative approach to generate event arguments end-to-end, and subsequently, PAIE \cite{PAIE} introduces more effective manually crafted prompts, using slot prompts to extract arguments by filling slots. TabEAE \cite{TabEAE} defines EAE as a table-filling problem, enabling the extraction of all events present in the input simultaneously. However, the aforementioned models did not simultaneously address capturing the semantic boundaries between events and As shown in the Figure \ref{fig:prefix}, CsEAE explicitly addresses both of the issues by incorporating co-occurrences- and structure-aware modules.

\subsection{Large Language Models for Event Argument Extraction}

The success of LLMs \cite{llama} has been widely recognized, and in recent years, there has been increasing research on the development of LLMs in the field of event extraction. Such as \cite{LLM-IE, LLM-EAE, LLM-star,LLM-EE-Annotator} have explored the performance of LLMs in event extraction tasks. However, these studies typically rely on In-context Learning (ICL). While this approach significantly conserves computational resources, it often results in less satisfactory outcomes compared to SLMs. In this paper, we move beyond the limitations of ICL and employ SFT, enabling LLMs to learn how to perform event extraction more effectively. We also found that multiple dataset SFT can improve the extraction capabilities of LLMs. Building on this, we introduced supplementary dataset enhancement training. Finally, we incorporated the insights derived from CsEAE into LLMs, achieving further improvements.

\section{CsEAE Model}

In this section, we will provide a detailed introduction to each component of CsEAE.
\subsection{Basic Architecture}

In the Figure \ref{fig:prefix}, given the input $\mathcal D$ and the prompt $p_{e_n}$ corresponding to the event type to be extracted, we fed $\mathcal D$ into an encoder with a structure-aware prefix, resulting in $H_{\mathcal D}^{enc}$. Then, $H_{\mathcal D}^{enc}$ is passed through a decoder with a co-occurrences-aware prefix to obtain the contextual representation of $\mathcal D$, referred to as the event-oriented context representation $H_{\mathcal D}$. This process can be formulated as:
\begin{equation}
        H_{\mathcal D}^{enc}=Encoder_{Sap}(\mathcal D),
    \label{eq:equation1}
\end{equation}
\begin{equation}
        H_{\mathcal D}=Decoder_{Cap}(H_{\mathcal D}^{enc},H_{\mathcal D}^{enc}).
    \label{eq:HD}
\end{equation}
Where $Sap$ represents structure-aware prefix, $Cap$ represents co-occurrences-aware prefix.

To create the span selector $\theta$, we need to interactively encode each token representation of $\mathcal D$ with $p_{e_n}$ at a deep level. Specifically, we will input $H_{\mathcal D}^{enc}$ and $p_{e_n}$ together into the Decoder after concatenating with the structure-aware prefix, obtaining its context-oriented prompt representation $H_{pt}$. We formalize it as:
\begin{equation}
        H_{pt}=Decoder_{Sap}(H_{\mathcal D}^{enc}, p_{e_n}).
    \label{eq:Hpt}
\end{equation}

\subsection{Co-occurrences-aware Module}

\label{sec:Co}


Co-occurrences-aware module introduces event co-occurrences-aware interaction through three aspects: context labeling, prompt extraction and co-occurrences prefix.

\subsubsection{Context Labeling} Given the input of the model $\mathcal D=\{t_{1},t_{2},\ldots,t_{n}\}$, where $t_i$ represents the $i$-th token in the input. Given $E=\{e_{0},e_{1},\ldots,e_{l}\}$, where $e_i$ represents one event appearing in $\mathcal D$, and $l$ represents the number of events appearing in $\mathcal D$. Given all the triggers $T=\{e_0^t,e_1^t,\ldots,e_l^t\}$, where $e_i^t$ represents the trigger corresponding to event $e_i$, and $e_i^t$ corresponds one-to-one with $e_i$. We will annotate all token spans corresponding to triggers in $\mathcal D$ according to the order in which the triggers $e_i^t$ appear in $\mathcal D$. Specifically, for the trigger $e_n^t$ corresponding to the event $e_n$ being extracted, we will annotate its appearance in $\mathcal D$ using special characters \textless t- -1\textgreater and \textless/t- -1\textgreater. 

For triggers $e_j^t$ corresponding to other events existing in $\mathcal D$, we will annotate them according to the order of appearance in $\mathcal D$ using \textless t-$k$\textgreater and \textless/t-$k$\textgreater, where $k$ is calculated starting from 0 and incremented by 1.

\subsubsection{Prompt Extraction} Given $P_{e}=\{p_{e_{1}},p_{e_{2}},\ldots,p_{e_{l}}\}$, where $p_{e_{i}}$ represents the prompt corresponding to event $e_i$. Notice that $p_{e_{i}}$, $e_i^t$ and $e_i$ are uniquely paired. In this paper, we utilize prompts proposed in PAIE \cite{PAIE} for the Rams and WikiEvents datasets and those in TabEAE \cite{TabEAE} for the MLEE dataset. To fully utilize the semantic information provided by the prompts, we first concatenate all prompts $P_e$ corresponding to events mentioned in $\mathcal D$. Then, we encode them into the SLMs to obtain dense vector representations $W_C$ for all co-occurring event prompts. Finally, the information of $W_C$ is integrated into the prefixes.

\subsubsection{Co-occurrences Prefix}
\label{sec: prefix}
After constructing the co-occurrences-aware matrix $W_C$ for the current event mention $\mathcal D$, we condense $W_C$ into prefixes \cite{PREFIX, AMPERE}, which then participate in the model's generation. As shown in the Figure \ref{fig:prefix}. Firstly, we introduce a learnable vector of length $len$, which serves as the Q vector for multi-head attention, where $len$ is a tunable hyperparameter controlling the final length of the prefixes to be fed into the SLMs, we set it as 40. Then, $W_C$ is used as the K and V vectors in multi-head attention computation, which is computed with the Q vector. After multi-head attention computation, we obtain a set of compressed dense vector $\mathcal P$, which then undergoes a series of linear layers. Finally, $\mathcal P$ is evenly split into $c$ segments ${\mathcal P}=\{\mathcal P_{1},\mathcal P_{2},\ldots,\mathcal P_{c}\}$, each with a length of $len$, where $c$ is the number of transformer layers in the SLMs. This results in prefixes that can be concatenated into the SLMs for computation.


\subsection{Structure-aware Module}
\label{sec:Structure}

structure-aware module introduces structure-aware interaction through two aspects: structural relationship and structure prefix.

\subsubsection{Structural Relationship} For different document inputs, as shown in Figure \ref{fig:prefix} (blue part on the right), we designed a structure-aware self-attention mask $M_s$, which treats sentences as units and trains the model to be structure-aware across the entire document. Specifically, given the document-level input $\mathcal D=\{S_{1},S_{2},\ldots,S_{m}\}$, where $S_i$ represents the $i$-th sentence in $\mathcal D$, and given the trigger $e_n^t$ of the current event to be extracted, located in sentence $S_n$, $M_s$ restricts the receptive field of all sentences except $S_n$, allowing these sentences to focus only on themselves and $S_n$. In contrast, $S_n$ can attend to all sentences. 

We can obtain the structure-aware dense vector representation $W_S$ for the event mention $\mathcal D$ as follows: 
\begin{equation}
        W_{S}=Decoder(Encoder(\mathcal D,M_{s})).
\end{equation}

\subsubsection{Structure Prefix} Finally, following the same approach as described in Section Co-occurrences-aware Prefix, the information from $W_S$ is integrated into the prefixes and participates in the model's generation.


\subsection{Span Selection}
\label{sec:spanselection}
After obtaining $H_{pt}$, we extract the slot representation $\psi_k$ corresponding to the pre-defined roles from $H_{pt}$, where $k$ represents the $k$-th slot. Then, we convert $\psi_k$ into a span selector specific to that slot $\theta_{k}$ \cite{PAIE,QAbased}. Next, apply the span selector $\theta_{k}$ directly to the event-oriented context representation $H_D$ to determine the argument's token span $[p_k^{(start)}; p_k^{(end)}]$.
\begin{equation}
\begin{gathered}
        \psi_{k}^{(start)}=\psi_{k}\circ w^{(start)}\in R^{h}, \\
        \psi_{k}^{(end)}=\psi_{k}\circ w^{(end)}\in R^{h}, \\
        \mathrm{logit}_k^{(start)}=\psi_k^{(start)}H_{\mathcal D}\in R^L, \\
        \mathrm{logit}_{k}^{(end)}=\psi_{k}^{(end)}H_{\mathcal D}\in R^{L}, \\
        p_k^{(start)}=\text{Softmax}(\text{logit}_k^{(start)})\in R^L, \\
        p_k^{(end)}=\text{Softmax}(\text{logit}_k^{(end)})\in R^L.
\end{gathered}
\label{eq:pk}
\end{equation}

Where $\theta=[w^{(start)};w^{(end)}] \in R^{h\times2}$ is a learnable parameter matrix shared by all span selectors, $\circ$ represents element-wise multiplication. $\theta_{k}=[\psi_{k}^{(start)};\psi_{k}^{(end)}]$ is the span selector specific to the slot corresponding to the role, $L$ demotes the context lengrth.

We define the loss function $\mathcal{L}$ as follows:
\begin{equation}
\begin{gathered}
        \mathcal{L}_k(\mathcal D)=-(\log p_k^{(start)}(s_k)+\log p_k^{(end)}(e_k)), \\
        \mathcal{L}=\sum_{\mathcal D\in B}\sum_k\mathcal{L}_k(\mathcal D).
\end{gathered}
\label{eq:l}
\end{equation}
Where $B$ ranges over all context in dataset and $k$ ranges over all slots in prompt $p_{e_{n}}$ for $\mathcal D$, and $(s_k, e_k)$ represents the token span of the most likely argument corresponding to the role in $H_D$.

During the inference phase, we predefine spans $\mathcal{C}$ that cover all possible spans within a predefined length and include a special span (0, 0) to represent the absence of any corresponding argument. Then, we utilize the span selector $\theta_{k}$  to compute scores for all spans using the following method:
\begin{equation}
       \mathrm{score}_k(i,j)=\mathrm{logit}_k^{(start)}(i)+\mathrm{logit}_k^{(end)}(j).
    \label{eq:score}
\end{equation}
Where $i$ and $j$ represent the start and end indices of each span in the set of spans.

Based on the scores, we determine the predicted final span by selecting the span with the highest score.
\begin{equation}
       (\widehat{s_k},\widehat{e_k})=\arg\max_{(i,j)\in\mathcal{C}}\text{score}_k(i,j).
    \label{eq:sk}
\end{equation}

For the issue of multiple arguments of the same role, we utilize the Hungarian algorithm \cite{Hungarian, PAIE}. For the problem of allocating multiple slots corresponding to a single role, we employ Bipartite Matching \cite{span_1, span_2, PAIE}.

\section{Generalization in LLMs}
In this section, we will provide a detailed explanation of how to use LLMs for EAE and further improvements.

\begin{figure}[!htbp]
    \centering
    \includegraphics[width=0.43\textwidth]{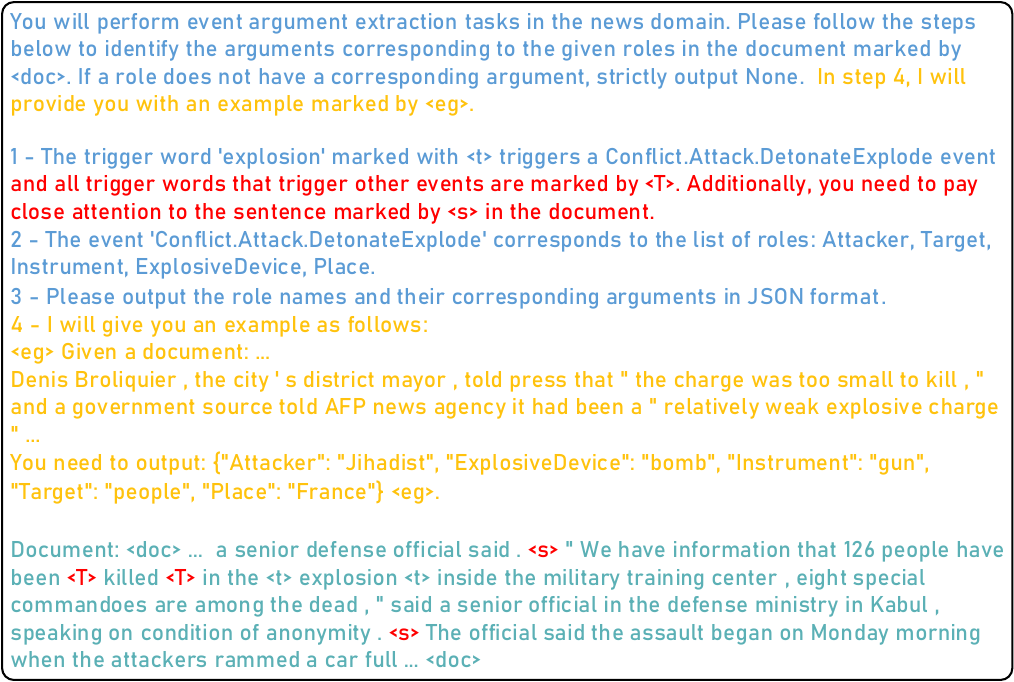}
    \caption{Prompt for LLMs on WikiEvents. The blue parts represent $\mathcal{I}$, the yellow parts represent $\mathcal{E}$, the green parts represent $\mathcal{Q}$ and the red parts represent co-occurrences- and structure-aware interactions.}
    \label{fig:CsLLM}
\end{figure}

\subsection{Prompt Design}
Given the input $\mathcal D$, we designed a corresponding prompt $\mathcal{P}_\mathcal{L}(\mathcal D)$ for LLMs. As shown in the Figure \ref{fig:CsLLM}, the prompt $\mathcal{P}_\mathcal{L}(\mathcal D)$ is divided into three parts:
\begin{equation}
       \mathcal{P}_{\mathcal{L}}(\mathcal D)=[\mathcal{I}; \mathcal{E}; \mathcal{Q}].
    \label{eq:equation2}
\end{equation}

The first part is the instruction $\mathcal{I}$, which describes the task and provides basic information such as the trigger, roles, and output format. The second part is the example $\mathcal{E}$, which provides a single example (one-shot) to the LLMs. We identified corresponding examples for each event type from the training set and the example should include as many arguments as possible from the input. The third part is the question $\mathcal{Q}$. We use \textless doc\textgreater for input to separate the $\mathcal{Q}$ from other components in the prompt.

\begin{table*}[!htbp]
    \centering
    \begin{tabular}{l|l|cccccc}
    \hline
        \multirow{2}{*}{Model} & \multirow{2}{*}{PLM} & \multicolumn{2}{c}{Rams} & \multicolumn{2}{c}{WikiEvents} & \multicolumn{2}{c}{MLEE} \\ \cline{3-8}
        ~ & ~ & Arg-I & Arg-C & Arg-I & Arg-C & Arg-I & Arg-C \\ \hline
        EEQA$^*$ & RoBERTa & 51.9 & 47.5 & 60.4 & 57.2 & 70.3 & 68.7 \\ 
        TSAR$^*$ & RoBERTa & 57.0 & 52.1 & \underline{71.1} & 65.8 & 72.6 & 71.5 \\
        BART-Gen$^*$ & BART & 51.2 & 47.1 & 66.8 & 62.4 & 71.0 & 69.8 \\ 
        TabEAE-m2s & RoBERTa & 56.2 & 51.4 & 69.7 & 64.9 & - & - \\ 
        TabEAE-m2m & RoBERTa & 55.9 & 50.9 & 70.3 & 64.6 & 74.0 & 72.9 \\ \hline
        PAIE & BART & 55.3 & 51.0 & 68.9 & 64.2 & 71.3 & 70.1 \\ \hline
        CsEAE & BART & \underline{57.5} & \underline{53.1} & 70.9 & \underline{66.5} & \underline{74.3} & \underline{73.3} \\ \hline
    \end{tabular}
    \caption{Overall performance of CsEAE and baselines. * means the value from the TabEAE's paper. All experiments utilized a large-scale PLM. The highest scores are underlined.}
    \label{tab:main}
\end{table*}

\subsection{Supervised Fine-Tuning}
SFT is the critical stage that endows the model with high-quality extraction capabilities. Through training data, the model can effectively leverage the latent knowledge accumulated during pre-training to understand and respond to extraction instructions \cite{zhongjing}.

A high-quality pre-training corpus can significantly enhance the performance of LLMs, even to the extent of breaking through scaling laws \cite{Pre-training-datasets}. Inspired by this, and considering the complexity of the EAE domain \cite{LLM-IE}, we sequentially merged multiple datasets and fine-tuned the LLMs using the combined dataset. To further exploit the improvements from multiple dataset SFT and enhance the model's sensitivity to extraction tasks, we incorporated additional datasets into the multiple dataset SFT, conducting enhanced training on the LLMs.

\subsection{CsLLMs}

In CsEAE, we optimized the model using event co-occurrences- and structure-aware interactions of the document. This brings up an important question: does insights that has been validated to be effective for extraction in SLMs also work effectively in LLMs? 

We believe this is a crucial question, as it can bridge future developments on LLMs with the extensive work previously done on SLMs. Therefore, we also incorporated event co-occurrences- and structure-aware interactions into the prompt. In the Figure \ref{fig:CsLLM}, the changes are highlighted in red. Specifically, we introduced co-occurrences-aware interaction in the $\mathcal{Q}$ by marking the triggers and introduced structure-aware interaction by marking the sentence containing the trigger. Additionally, we guided the model in the $\mathcal{I}$ to pay attention to these marked pieces of information. We refer to the fine-tuned LLMs, which integrate the information mentioned above, as CsLLMs.

\section{Experiments}

\subsection{Experimental Setup}

\subsubsection{Datasets} We used the three most commonly employed datasets for document-level event argument extraction (EAE): Rams \cite{RAMS}, WikiEvents \cite{DOC_EE}, and MLEE \cite{MLEE}. We preprocessed the data following previous methods \cite{MLEE-process,PAIE,TabEAE}. To further enhance model training, we also incorporated sentence-level EAE datasets, specifically ACE \cite{ACE} and GENEVA \cite{geneva}, applying preprocessing techniques from prior research \cite{DEGREE,AMPERE,geneva}.

Additionally, to more comprehensively validate the effectiveness of CsEAE, we applied the data processing methods used in TextEE~\cite{TextEE} to WikiEvents and RAMS. These methods included standardization of data assumptions, normalization of data processing steps, and standardization of dataset splits (5 times).

\subsubsection{Implementation Details} We used PyTorch and a single NVIDIA A40 Tensor Core GPU with 45GB to train all models and reproduce experiments of other models. We used BART \cite{BART} as the backbone for CsEAE. During model training the learning rate was set to 2e-5. We used the methods provided by LLama-Factory \footnote{https://github.com/hiyouga/LLaMA-Factory} for model's SFT, employing LoRA-based \cite{LLM-lora} fine-tuning with a rank r of 8 and a dropout rate of 0.1. The batch size was set to 4, and training was conducted for 3 epochs on each dataset.

\subsubsection{Evaluation Metrics} Fllowed by previous works \cite{PAIE,TabEAE}, We used the Arg-I F1 and Arg-C F1 metrics to evaluate the model's performance on the argument identification and argument classification.

It should be noted that our use of Arg-I and Arg-C corresponds to  Arg-I+ and Arg-C+ as defined in TextEE, meaning that predicted arguments are attached to the correct trigger words. In all experiments in this paper,  Arg-I and Arg-C is equivalent to  Arg-I+ and Arg-C+.

\subsubsection{Baselines} For SLMs, we categorized the baseline models into two groups: 

(1) Classification-based models: EEQA \cite{EEQA}, TSAR \cite{TSAR}, TagPrime-C and TagPrime-CR~\cite{TAGPRIME}; 

(2) Generation-based models: Bart-Gen \cite{DOC_EE}, X-Gear \cite{X-gear}, AMPERE \cite{AMPERE}, PAIE \cite{PAIE}, TabEAE \cite{TabEAE}. 

For LLMs, we categorized the baseline models into two groups too: 

(1) Open-AI: Chat-GPT \footnote{The versions of model we use are: gpt-3.5-turbo-0125}, GPT-4o, GPT-4o-mini; 

(2) Open-source: Llama3-8B~\footnote{https://huggingface.co/meta-llama/Meta-Llama-3-8B}~\cite{llama}, Llama3-8B-Instruct \footnote{https://huggingface.co/meta-llama/Meta-Llama-3-8B-Instruct}.

\begin{table*}[!h]
    \centering
    \begin{tabular}{l|l|cccc}
    \hline
        \multirow{2}{*}{Model} & \multirow{2}{*}{PLM} & \multicolumn{2}{c}{Rams} & \multicolumn{2}{c}{WikiEvents}  \\ \cline{3-6}
        ~ & ~  & Arg-I & Arg-C & Arg-I & Arg-C   \\ \hline
        TagPrime-C$^*$ & RoBERTa & 54.4 & 48.3 &  68.6 & 64.0 \\
        TagPrime-CR$^*$ & RoBERTa & 54.1 & 49.7 & 68.4 & 65.5   \\
        EEQA$^*$ & BART & 48.9 & 44.7 & 48.4 & 46.1  \\ 
        BART-Gen$^*$ & BART & 50.4 & 45.4 & 68.1 & 63.9  \\ 
        X-Gear$^*$ & BART & 52.1 & 46.2 & 55.4 & 52.4  \\ 
        Ampere$^*$ & BART & 52.0 & 46.8 & 56.2 & 53.3  \\ \hline
        PAIE & BART & 56.4 & 51.9 & 68.5 & 64.5  \\ \hline
        CsEAE & BART & \underline{56.8} & \underline{52.3} & \underline{69.3} & \underline{65.7} \\ \hline
    \end{tabular}
    \caption{All experiments in the table above used the data processing methods described in TextEE, and the results are averaged over five data splits. * means the value from the TextEE's paper. All experiments utilized a large-scale PLM.}
    \label{tab:main-textee}
\end{table*}

\subsection{Main Results}


\begin{table*}[!htbp]
    \centering
    \begin{tabular}{l|llllll}
    \hline
        \multirow{2}{*}{Model}  & \multicolumn{2}{c}{WikiEvents}& \multicolumn{2}{c}{Rams} & \multicolumn{2}{c}{MLEE}   \\ \cline{2-7}
        ~ & Arg-I & Arg-C & Arg-I & Arg-C & Arg-I & Arg-C  \\ \hline
        \multicolumn{7}{l}{In-Context Learning (ICL)} \\ \hline
        GPT-3.5 & 18.12 & 16.04 & 34.30 & 27.64 & 21.16 & 15.46  \\ 
        GPT4o-mini  & 20.42 & 17.99 & 35.47 & 30.04 & 25,85 & 22.34  \\ 
        GPT4o  & 25.58 & 23.37 & 41.58 & 35.70 & 28.04 & 24.92 \\ 
        Llama3  & 10.34 & 9.50 & 23.05 & 18.79 & 0.07 & 0.07  \\ 
        Llama3-Instruct  & 0.00 & 0.00 & 0.00 & 0.00 & 0.00 & 0.00  \\ \hline
        \multicolumn{7}{l}{Supervised Fine-tuning} \\ \hline
        Llama3  & 65.82 & 60.68 & 37.00 & 33.26 & 72.63 & 71.09  \\ 
        Llama3-Instruct  & 65.88 & 60.54 & 55.06 & 49.82 & 70.85 & 69.76  \\
        CsLLMs  & 66.33 & 62.80 & 55.35 & 50.25 & 74.80 & 73.87   \\ \hline
        \multicolumn{7}{l}{Multiple Datasets Supervised Fine-tuning} \\ \hline
        Doc  & 66.73 & 62.99 & 55.76 & 50.74 & 73.35 & 71.96  \\
        CsLLMs (Doc)  & 69.92 & 65.66 & 56.14 & 50.99 & 75.34 & 74.10 \\ \hline
        \multicolumn{7}{l}{Multiple Datasets Supervised Fine-tuning using additional datasets} \\ \hline
        News  & 69.12 & 63.70 & 56.12 & 51.32 & - & -  \\
        News+MLEE  & 68.92 & 65.12 & 54.96 & 50.82 & 70.83 & 69.58 \\ 
        News+GENEVA  & 57.85 & 54.54 & 44.12 & 41.04 & - & -  \\ 
        ALL  & 68.27 & 63.96 & 56.83 & 51.62 & 72.04 & 70.87 \\ \hline
        CsLLMs (ALL)  & \underline{70.89} & \underline{66.53} & \underline{57.19} & \underline{51.84} & \underline{75.93} & \underline{74.89}   \\ \hline
    \end{tabular}
    \caption{Overall performance of LLMs. Doc represents training using the WikiEvents, Rams, and MLEE; News represents training using the ACE, Rams, and WikiEvents, ALL signifies that all five datasets were used for training. CsLLMs refers to the fine-tuning process that incorporates prompts enhanced with co-occurrences- and structure-aware interactions. During the multiple dataset SFT, we used Llama3-Instruct as the LLMs.}

    \label{tab:llm-main}
\end{table*}

\subsubsection{CsEAE}

We evaluate the proposed model CsEAE and baseline methods under all benchmarks. In the Table \ref{tab:main}, our model outperformed all baselines on the Rams and MLEE datasets.

\begin{table}[!ht]
    \centering
    \begin{tabular}{l|cc}
    \hline
         \multicolumn{3}{c}{GENEVA}  \\ \hline
        Model & Arg-I & Arg-C   \\ \hline
        \multicolumn{3}{l}{In-Context Learning (ICL)} \\ \hline
        GPT-3.5 & 33.07 & 27.97 \\
        GPT4o-mini & 35.17 & 31.06   \\
        GPT4o & 42.98 & 39.55   \\ 
        Llama3 & 4.70 & 3.61   \\ 
        Llama3-Instruct & 0.35 & 0.29   \\ \hline
        \multicolumn{3}{l}{Supervised Fine-tuning} \\ \hline
        Llama3 & 28.98 & 27.88   \\ 
        Llama3-Instruct & 66.07 & 62.42   \\ 
        News+GENEVA & 64.22 & 61.06   \\ 
        ALL & 63.91 & 61.03  \\ \hline
        CsLLMs (ALL) & \underline{67.99} & \underline{64.71}  \\ \hline
    \end{tabular}
    \caption{Overall performance of LLMs on GENEVA.}
    \label{tab:llm-geneva}
\end{table}

Compared to the baseline model PAIE \cite{PAIE}, CsEAE achieves improvements on the Rams dataset, with increases of 2.2\% and 2.1\%, respectively. On the WikiEvents dataset, CsEAE shows improvements of 2.0\% in Arg-I and 2.3\% in Arg-C metrics. Similarly, on the MLEE dataset, CsEAE achieves improvements of 3.0\% in Arg-I and 3.2\% in Arg-C metrics. The consistent improvements of 2\% or more across all datasets demonstrate the effectiveness of the structure- and co-occurrences-aware modules in document-level EAE tasks.

We also utilized the data preprocessing method provided by TextEE, dividing the dataset into five subsets while allowing for multi-word triggers, accounting for overlapping argument spans, and retaining all instances without filtering. The final results, shown in the Table~\ref{tab:main-textee}, represent the average performance across these five splits. Even under such stringent conditions, CsEAE consistently outperforms all baselines, demonstrating its superior effectiveness.
\subsubsection{LLMs}

As shown in the Table \ref{tab:llm-main}, in the ICL setting, the Open-AI series models demonstrated superior performance compared to the Open-resource models. Notably, instruct-type models have shown relatively poor performance during ICL. However, after fine-tuning, they outperformed base models on some datasets.

\begin{table*}[ht]
    \centering
    \begin{tabular}{l|cccccc}
    \hline
        \multirow{2}{*}{Model} & \multicolumn{2}{c}{Rams} & \multicolumn{2}{c}{WikiEvents} & \multicolumn{2}{c}{MLEE} \\ \cline{2-7}
        ~ & Arg-I & Arg-C & Arg-I & Arg-C & Arg-I & Arg-C \\ \hline
        w/o str\&occur & 55.3 & 51.0 & 68.9 & 64.2 & 71.3 & 70.1 \\ \hline
        add str & 55.8 & 52.0 & 70.5 & 64.8 & 72.0 & 70.9 \\
        add occur & 55.9 & 51.6 & 70.5 & 65.9 & 73.9 & 72.9 \\
        CsEAE & \underline{57.5} & \underline{53.1} & \underline{70.9} & \underline{66.5} & \underline{74.3} & \underline{73.3} \\ \hline
    \end{tabular}
    \caption{Ablation study on all benchmarks, str: structure-aware interaction, occur: co-occurrences-aware interaction.}
    \label{tab:ablationstudy}aai25
\end{table*}

\begin{table*}[!htbp]
    \centering
    \begin{tabular}{l|cccc|cccc}
    \hline
        \multirow{4}{*}{Model} & \multicolumn{4}{c}{WikiEvents} & \multicolumn{4}{c}{MLEE} \\ \cline{2-9}
        ~ & \multicolumn{2}{c|}{N\_O (296)} & \multicolumn{2}{c|}{Overlap (69)} & \multicolumn{2}{c|}{N\_O (734)} & \multicolumn{2}{c}{Overlap (1460)} \\ 
        ~ & Arg-I & \multicolumn{1}{c|}{Arg-C} & Arg-I & \multicolumn{1}{c|}{Arg-C} & Arg-I & \multicolumn{1}{c|}{Arg-C} & Arg-I & Arg-C \\ \hline
        TabEAE & 70.7 & 65.4 & 66.1 & 63.0 & 78.0 & 77.0 & 68.9 & 67.6 \\ 
        PAIE & 68.8 & 63.9 & 68.9 & 65.0 & 76.8 & 75.7 & 64.8 & 63.4 \\ 
        CsEAE & \underline{71.0} & \underline{66.0} & \underline{70.6} & \underline{68.4} & \underline{78.7} & \underline{77.8} & \underline{69.0} & \underline{67.9} \\ \hline
    \end{tabular}
    \caption{The performance in extracting the arguments of overlapping events. The numbers in parentheses represent the quantity of the corresponding data type within the dataset.}
    \label{tab:overlap}
\end{table*}

After SFT, the extraction capabilities of the LLMs improved significantly. Further improvements were observed when the model was fine-tuned using multiple datasets, demonstrating that the LLMs robust memory capacity can handle diverse datasets simultaneously and learn common extraction-enhancing abilities from them. Additionally, after incorporating two extra sentence-level datasets for enhanced training, the model achieved better performance.

Moreover, incorporating co-occurrences- and structure-aware interactions into the prompts led to additional performance gains compared to models fine-tuned on single datasets without such enhancements. This indicates that beneficial extraction-related insights identified in SLMs is also applicable and effective in LLMs.

We attribute the lower performance of CsLLMs (ALL) on Rams compared to CsEAE to the incomplete integration of structure-aware elements in the prompt. While structure-aware interaction has been proven to be the most effective module for improving Rams performance in CsEAE (analysis on ablation studies), but we are unable to fully constrain the model's focus through the prompt alone.

To analyze the generalization challenges of LLMs in broader domains and their applicability in real-world scenarios, we conducted extensive experiments on the GENEVA dataset, which includes 115 event types and 220 distinct roles across general-domain, sentence-level data. The experimental results are presented in the table~\ref{tab:llm-geneva}. Surprisingly, unlike in domain-specific document-level datasets,  multiple datasets SFT does not enhance model performance on GENEVA. However, incorporating co-occurrences- and structure-aware interactions into the prompt improves the model's performance on document-level datasets, allowing for better extraction on GENEVA. This indicates that the model learns to capture co-occurrences- and structure-aware information from the three document-level datasets, such that, even though sentence-level datasets cannot directly embed structure-aware information in prompt construction, the model can leverage what it learned from document-level data to assist in extraction.
Additionally, it becomes evident that LLMs do not perform well on general-domain datasets like GENEVA. Its best performance, an Arg-C score of 64.71, falls short compared to best results of SLMs ~\cite{TextEE}. We attribute this to the fact that many event types in GENEVA are quite similar, and fine-tuning an 8B-parameter model using prompt + LoRA struggles to discern numerous labels and their subtle interactions during extraction~\cite{LLM-IE}.
\section{Analysis}

\subsection{Ablation Studies}

The Table \ref{tab:ablationstudy} show that even a single type of interaction can enhance the model's performance across all datasets, with each interaction type providing varying levels of improvement. The structure-aware module significantly improves performance on the Rams dataset, increasing the Arg-C metric by 1\%. Conversely, the co-occurrence-aware module significantly boosts performance on the WikiEvents and MLEE datasets, increasing the Arg-C metric by 1.7\% and 2.8\%, respectively. We analyzed that the significant improvement in Rams by structure-aware module is due to its stable sentence structure, where each document consists of five sentences, allowing the model to learn more consistent structural information. The notable improvement of the co-occurrences-aware module on the WikiEvents and MLEE datasets is attributed to the higher number of events in instances, where the auxiliary information provided by the co-occurrences-aware module leads to a greater performance boost in complex event scenarios. CsEAE not only retains the benefits of individual interaction features but also integrates multiple types of interaction without causing interference.


\subsection{Capturing the Event Semantic Boundary}
\label{sec:Boundary}

Following TabEAE, we analyzed CsEAE's ability to capture event semantic boundaries on the WikiEvents and MLEE datasets from two perspectives: inter-event semantics and intra-event semantics.

\subsubsection{Inter-event semantics} We divided the both datasets based on the overlap, where overlap indicates instances where different events use the same token span as arguments, and N\_O denotes instances without event overlap. As observed from the Table \ref{tab:overlap}, CsEAE achieved overall improvements across all metrics on both datasets and performed particularly well in handling instances with overlap.

\subsubsection{Inner-event semantics} 
\begin{figure}[!htbp]
    \begin{minipage}[t]{0.48\linewidth}
        \centering
        \includegraphics[width=\textwidth]{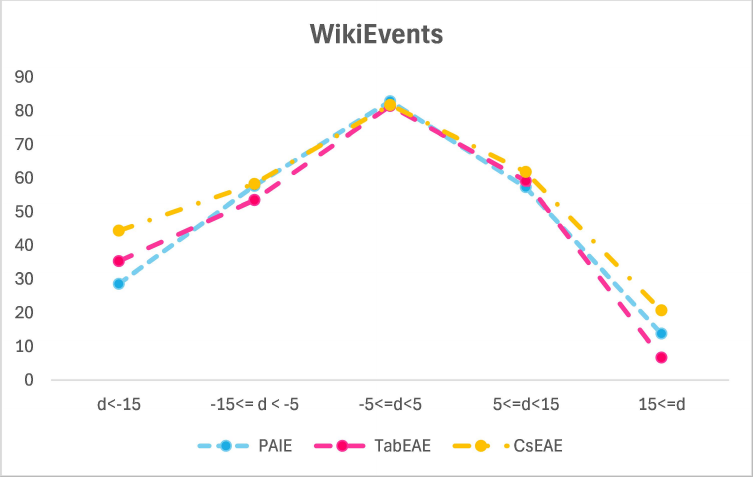}
    \end{minipage}%
    \begin{minipage}[t]{0.5\linewidth}
        \centering
        \includegraphics[width=\textwidth]{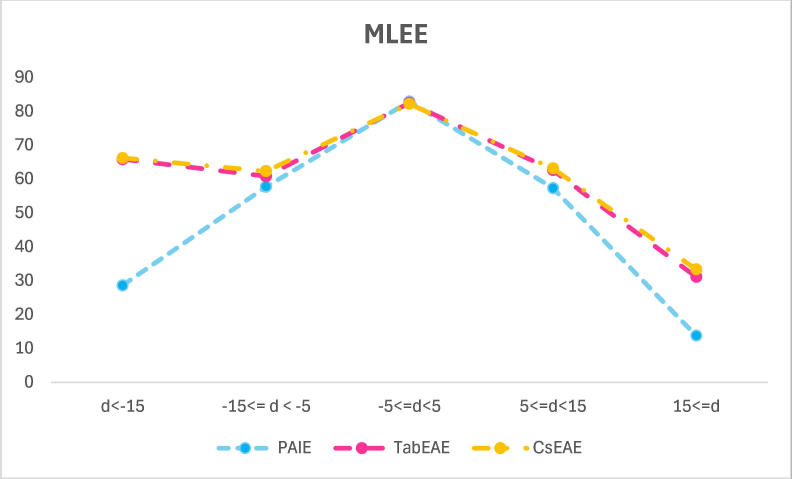}
    \end{minipage}
    \caption{The performance of different EAE models in extracting arguments at different distances from the triggers. We only measure the Arg-C F1 metric.}
    \label{fig:d}
\end{figure}

We divided the roles in the both datasets based on their distance from the trigger. Specifically, we defined the argument distance as the value obtained by subtracting the index of the argument's head word from the index of its corresponding trigger's head word. Since the model predicts all arguments corresponding to a role at once, we defined the distance between a role and the trigger, $\mathcal D$, as the maximum argument distance among all arguments for that role. As shown in the Figure \ref{fig:d}, where negative values indicate the argument is to the left of the trigger and positive values indicate the argument is to the right. The results show that CsEAE achieved the best performance across multiple ranges on both datasets and demonstrated a trend where the improvement increased with greater distances.

\subsection{Structure-aware Interaction for Document}
\label{sec: strufordoc}

To analyze the effectiveness of the model in performing extraction centered around the sentence containing the trigger word, we conducted an analysis on RAMS, which has the highest number of cross-sentence arguments.
We defined the distance D between a role and the trigger as the maximum argument distance among all arguments for that role. When the trigger and the maximum argument are in the same sentence, D=0; when they are not, D$\neq$0. In the Table \ref{tab:D=0}, CsEAE achieved a 3.23\% improvement in the Arg-C metric compared to PAIE when D=0. This improvement significantly contributed to CsEAE's overall lead over PAIE in all datasets. The substantial improvement at D=0 also demonstrates that the model's approach of centering the document structure around the trigger's sentence effectively helps focus attention on the core content of the sentence, reducing the distraction from redundant information. 

\begin{table}[!htbp]
    \centering
    \begin{tabular}{l|ccc}
    \hline
        \multirow{2}{*}{Model} & \multicolumn{3}{c}{Rams (Arg-C F1)} \\ \cline{2-4}
        ~ & D=0 & D$\neq$0 & Overall \\ \hline
        PAIE & 58.7 & 35.3 & 51.0 \\
        TabEAE & 61.2 & 31.8 & 51.4 \\
        CsEAE & \underline{61.9} & \underline{35.5} & \underline{53.1} \\ \hline
    \end{tabular}
    \caption{Model performance on cross-sentence arguments.}
    \label{tab:D=0}
\end{table}

\subsection{Case Study}

\begin{figure}[!htbp]
    \centering
    \includegraphics[width=0.4\textwidth]{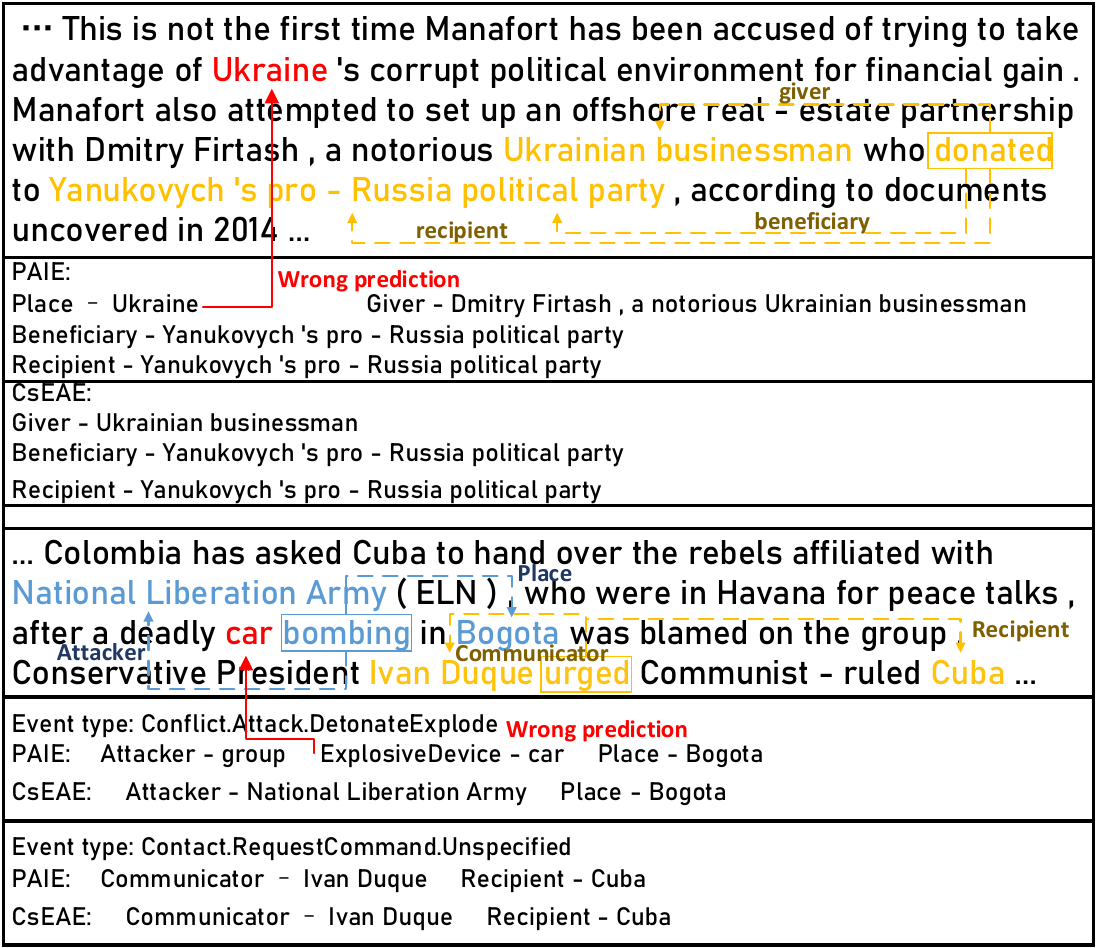}
    \caption{Two test cases from Rams and WikiEvents.}
    \label{fig:case}
\end{figure}

In the first case in the Figure \ref{fig:case}, PAIE incorrectly predicts \textit{Ukraine} from the previous sentence as the argument for role \textit{Place}, while CsEAE avoids this interference. In the second example, PAIE incorrectly identifies \textit{car} as the argument for \textit{ExplosiveDevice}, whereas CsEAE, by incorporating more event information, avoids this mistake.

\section{Conclusion}

We proposed CsEAE, which aids the model in capturing semantic boundaries between events by incorporating interaction of all co-occurring events within the input. Additionally, it establishes the structural relationships within the input to reduce the distraction caused by redundant information. Moreover, we addressed the gap in previous research regarding fine-tuning LLMs in the EAE domain, achieving further improvements through multiple dataset fine-tuning. Lastly, we intend to apply the reliable insights developed in CsEAE to LLMs, offering a new perspective: reliable insights validated on SLMs are also applicable to LLMs.

\section{Acknowledgement}
The article is supported by the “Tianshan Talent” Research Project of Xinjiang (No.2022TSYCLJ0037). The National Natural Science Foundation of China (No.62262065). The Science and Technology Program of Xinjiang (No.2022B01008).

Thanks to Professor Rui Yan (Renmin University) for the support and assistance, and to Haokun Geng and Guanghan Li (Xinjiang University) for their support.

Thanks to all the reviewers for their valuable feedback and to all the contributors who continuously advance the development of open-source projects in the EE community.

\bibliography{aaai25}

\section*{Reproducibility Checklist}


\subsection*{This paper:}

\begin{enumerate}
    \item Includes a conceptual outline and/or pseudocode description of AI methods introduced (yes)
    \item Clearly delineates statements that are opinions, hypothesis, and speculation from objective facts and results (yes)
    \item Provides well marked pedagogical references for less-familiar readers to gain background necessary to replicate the paper (yes)
\end{enumerate}

\subsection*{Does this paper make theoretical contributions? (yes)}

If yes, please complete the list below.

\begin{enumerate}
    \item All assumptions and restrictions are stated clearly and formally. (yes)
    \item All novel claims are stated formally (e.g., in theorem statements). (yes)
    \item Proofs of all novel claims are included. (yes)
    \item Proof sketches or intuitions are given for complex and/or novel results. (yes)
    \item Appropriate citations to theoretical tools used are given. (yes)
    \item All theoretical claims are demonstrated empirically to hold. (yes)
    \item All experimental code used to eliminate or disprove claims is included. (yes)
\end{enumerate}

\subsection*{Does this paper rely on one or more datasets? (yes)}

If yes, please complete the list below.

\begin{enumerate}
    \item A motivation is given for why the experiments are conducted on the selected datasets (yes)
    \item All novel datasets introduced in this paper are included in a data appendix. (yes)
    \item All novel datasets introduced in this paper will be made publicly available upon publication of the paper with a license that allows free usage for research purposes. (NA)
    \item All datasets drawn from the existing literature (potentially including authors’ own previously published work) are accompanied by appropriate citations. (yes)
    \item All datasets drawn from the existing literature (potentially including authors’ own previously published work) are publicly available. (yes)
    \item All datasets that are not publicly available are described in detail, with explanation why publicly available alternatives are not scientifically satisficing. (NA)
\end{enumerate}

\subsection*{Does this paper include computational experiments? (yes)}

If yes, please complete the list below.

\begin{enumerate}
    \item Any code required for pre-processing data is included in the appendix. (yes)
    \item All source code required for conducting and analyzing the experiments is included in a code appendix. (yes)
    \item All source code required for conducting and analyzing the experiments will be made publicly available upon publication of the paper with a license that allows free usage for research purposes. (yes)
    \item All source code implementing new methods have comments detailing the implementation, with references to the paper where each step comes from (yes)
    \item If an algorithm depends on randomness, then the method used for setting seeds is described in a way sufficient to allow replication of results. (NA)
    \item This paper specifies the computing infrastructure used for running experiments (hardware and software), including GPU/CPU models; amount of memory; operating system; names and versions of relevant software libraries and frameworks. (yes)
    \item This paper formally describes evaluation metrics used and explains the motivation for choosing these metrics. (yes)
    \item This paper states the number of algorithm runs used to compute each reported result. (yes)
    \item Analysis of experiments goes beyond single-dimensional summaries of performance (e.g., average; median) to include measures of variation, confidence, or other distributional information. (yes)
    \item The significance of any improvement or decrease in performance is judged using appropriate statistical tests (e.g., Wilcoxon signed-rank). (yes)
    \item This paper lists all final (hyper-)parameters used for each model/algorithm in the paper’s experiments. (yes)
    \item This paper states the number and range of values tried per (hyper-)parameter during development of the paper, along with the criterion used for selecting the final parameter setting. (yes)
\end{enumerate}

\end{document}